\pdfoutput=1
\documentclass[11pt]{article}

\usepackage{ACL2023}
\usepackage{times}
\usepackage{latexsym}

\usepackage[T1]{fontenc}
\usepackage[utf8]{inputenc}

\usepackage{microtype}

\usepackage{inconsolata}
\usepackage{url}

\usepackage{amsmath,amsfonts,bm}

\newcommand{\eg}{{\em e.g.,}}

\def\eqref#1{equation~\ref{#1}}
\def\1{\bm{1}}

\DeclareMathAlphabet{\mathsfit}{\encodingdefault}{\sfdefault}{m}{sl}
\SetMathAlphabet{\mathsfit}{bold}{\encodingdefault}{\sfdefault}{bx}{n}

\usepackage{amsmath}
\usepackage{multirow}
\usepackage{xcolor}

\usepackage{xcolor,colortbl}
\definecolor{lightblue}{rgb}{0.68, 0.85, 0.9}
\definecolor{lavender}{rgb}{0.9, 0.9, 0.98}
\definecolor{lightyellow}{rgb}{1.0, 1.0, 0.88}
\definecolor{magicmint}{rgb}{0.67, 0.94, 0.82}
\definecolor{palepink}{rgb}{0.98, 0.85, 0.87}
\definecolor{bubbles}{rgb}{0.91, 1.0, 1.0}

\definecolor{hidden-red}{RGB}{205, 44, 36}
\definecolor{hidden-blue}{RGB}{228,245,252}
\definecolor{hidden-orange}{RGB}{243,202,120}
\definecolor{hidden-green}{RGB}{34,139,34}
\definecolor{hidden-pink}{RGB}{255,245,247}
\definecolor{hidden-black}{RGB}{20,68,106}

\usepackage{microtype}
\usepackage{graphicx}
\usepackage{subfigure}
\usepackage{amssymb}
\usepackage{CJKutf8}
\usepackage{algorithm}
\usepackage{algorithmic}
\usepackage{enumitem}
\algsetup{linenosize=\small}
\usepackage{wrapfig}
\usepackage{cleveref}
\usepackage{multicol}
\usepackage{booktabs}
\usepackage{xspace}
\usepackage{adjustbox}
\usepackage{subfiles}
\usepackage{latexsym}

\usepackage{placeins}
\usepackage[french,vietnamese,main=english]{babel}
\usepackage{tikz}
\usepackage{tikz-qtree}
\usetikzlibrary{bayesnet}
\usetikzlibrary{trees,shapes,arrows}
\usepackage[switch]{lineno}
\usepackage[edges]{forest}
\usetikzlibrary{shapes, arrows.meta, positioning, shadows}
\babelprovide[import]{thai}

\newcommand*{\affaddr}[1]{#1} % No op here. Customize it for different styles.
\newcommand*{\affmark}[1][*]{\textsuperscript{#1}}
\newcommand*{\email}[1]{\textrm{#1}}

\definecolor{hidden-black}{RGB}{64,64,64}

\title{Data Augmentation using Large Language Models: \\Data Perspectives, Learning Paradigms and Challenges}

\author{Bosheng Ding\thanks{\; Equal contribution, order decided by coin flip.}\affmark[~~1]~~Chengwei Qin\footnotemark[1]\affmark[~~1]~~Ruochen Zhao\footnotemark[1]\affmark[~~1]~~\textbf{
Tianze Luo\affmark[1]~~}~~\textbf{Xinze Li\affmark[1]}\\\textbf{
Guizhen Chen\affmark[1]~~}\textbf{
Wenhan Xia\affmark[2]~~}
\textbf{Junjie Hu\affmark[3]}~~
\textbf{Anh Tuan Luu\affmark[1]}~~ \textbf{
Shafiq Joty\affmark[1,4]~~}\\
\affaddr{\affmark[1]Nanyang Technological University, Singapore}
\affaddr{\affmark[2]Princeton University}\\
\affaddr{\affmark[3]University of Wisconsin–Madison}
\affaddr{\affmark[4]Salesforce AI}\\
\email{\small{\{bosheng001, chengwei003, ruochen002, tianze001, xinze002, guizhen001, anhtuan.luu\}@ntu.edu.sg;}}\\
\email{\small{wxia@princeton.edu;}}
\email{\small {junjie.hu@wisc.edu;}}
\email{\small {sjoty@salesforce.com}} }

\begin{document}
 \maketitle
\begin{abstract}
In the rapidly evolving field of large language models (LLMs), data augmentation (DA) has emerged as a pivotal technique for enhancing model performance by diversifying training examples without the need for additional data collection. This survey explores the transformative impact of LLMs on DA, particularly addressing the unique challenges and opportunities they present in the context of natural language processing (NLP) and beyond.  From both data and learning perspectives, we examine various strategies that utilize LLMs for data augmentation, including a novel exploration of learning paradigms where LLM-generated data is used for diverse forms of further training.  Additionally, this paper highlights the primary open challenges faced in this domain, ranging from controllable data augmentation to multi-modal data augmentation. This survey highlights a paradigm shift introduced by LLMs in DA, and aims to serve as a comprehensive guide for researchers and practitioners.

% This survey paper presents an exhaustive exploration of data augmentation techniques in Natural Language Processing (NLP), highlighting their transformative impact on model performance and robustness. It delves into the myriad of methods employed for augmenting textual data, including text rewriting, synonym replacement, advanced generative models, and noise injection techniques. The paper systematically categorizes and evaluates these methods, offering insights into their applications across various NLP tasks such as text classification, language translation, and sentiment analysis. Furthermore, it discusses the critical role of evaluation metrics and benchmarks in determining the efficacy of augmentation strategies. Ethical considerations, particularly in terms of bias identification and mitigation, are emphasized to underscore the importance of responsible AI practices. Emerging trends and future directions in data augmentation, driven by advancements in AI and machine learning, are also explored, providing a comprehensive overview of the state-of-the-art and a glimpse into the future of NLP. This survey aims to serve as a valuable resource for researchers and practitioners alike, offering a thorough understanding of the current landscape and future prospects in NLP data augmentation.

\end{abstract}

% 1. Introduction
% 1.1 Background and Importance of the Study
% 1.2 Aim and Objectives of the Paper
% 1.3 Organization of the Paper

% 2. Background
% 2.1 Introduction to Artificial Intelligence
% 2.2 Overview of Health Care System
% 2.3 Need for AI in Health Care

% 3. Literature Review
% 3.1 Previous Surveys and Studies on AI in Health Care
% 3.2 Gaps in the Current Literature
% 3.3 Methodology for This Survey

% 4. Application of AI in Health Care
% 4.1 Diagnostic AI Systems
% 4.2 Predictive Analytics
% 4.3 AI in Precision Medicine
% 4.4 AI in Medical Imaging
% 4.5 AI in Patient Monitoring and Care
% 4.6 AI in Drug Discovery and Development
% 4.7 AI in Public Health Surveillance

% 5. Emerging Trends and Technologies
% 5.1 AI-enabled Medical Devices and Wearables
% 5.2 AI and Telemedicine
% 5.3 AI and Robotics in Health Care
% 5.4 AI in Genome Sequencing
% 5.5 AI in Personalized Medicine and Digital Therapeutics

% 6. Challenges and Barriers in Implementation
% 6.1 Technical Challenges
% 6.2 Ethical and Legal Issues
% 6.3 Data Privacy and Security Concerns
% 6.4 Challenges in Integration with Existing Systems
% 6.5 Need for Skilled Personnel

% 7. Case Studies
% 7.1 Success Stories of AI Implementation in Health Care
% 7.2 Lessons Learned

% 8. Future Directions
% 8.1 Predicted Advancements in AI for Health Care
% 8.2 Policy Recommendations for the Implementation of AI in Health Care

% 9. Conclusion
% 9.1 Summary of Findings
% 9.2 Implications for Practice and Policy
% 9.3 Suggestions for Future Research

\section{Introduction}
% \textbf{现在用LLM生成数据越来越重要，因为人类真实数据可能很快要用完了。我们不可避免要用synthetic data去做}
% \textbf{新的大模型用来生成数据比以往有很大的进步，打开了很大的空间，可以达到human level的水平}
% \textbf{LLM有新的learning 比如alignment、instruction-tuning，所以会有新的learning paradigm}
% \textbf{Challenges and Future Directions. Summarize 一下放进来}
% Data augmentation (DA) emerges as a cornerstone in the domain of machine learning (ML), aimed at tackling one of the most significant challenges: the scarcity of training data. 

Data-centric approaches to Artificial Intelligence (AI) constitute a pivotal element in the advancement towards Artificial General Intelligence (AGI), centering on the construction of AI systems underpinned by high-quality data \cite{zha2023data}. This emphasis on data quality is vital, as it ensures the clarity of the information from which AI systems are to learn. Nevertheless, the acquisition of high-quality data presents significant challenges, being both costly and time-intensive, while the data annotation phase is often laborious and prone to inaccuracies resulting from human involvement \cite{ding2022gpt}. In response to these challenges, researchers have dedicated efforts towards data augmentation (DA) techniques as a means to mitigate such issues \cite{chen-etal-2023-empirical}. Data augmentation fundamentally involves the adoption of innovative methods aimed at bolstering model efficacy through the broadening of training data diversity, all without necessitating further data collection efforts. This strategy effectively tackles a major hindrance in machine learning research, namely the dearth of readily accessible training data, by applying slight modifications to existing datasets or creating synthetic data instances.

% Data augmentation (DA) emerges as a cornerstone in the domain of machine learning (ML) for enhancing model performance \cite{feng-etal-2021-survey}. At its core, DA embodies innovative strategies to enhance model performance by increasing the diversity of training examples without the need for additional data collection \cite{chen-etal-2023-empirical}. This addresses a critical obstacle in ML research — the scarcity of available training data — by either introducing minor modifications to existing data or generating synthetic examples \cite{ding-etal-2020-daga, liu-etal-2021-mulda}. Such methodologies are essential in enriching the training dataset, thereby not only enhancing the robustness of models but also significantly reducing the likelihood of overfitting \cite{zhang2017mixup}. This approach serves as a crucial regularizer within the deep learning paradigm, particularly as models become increasingly complex and data-hungry \cite{cubuk2018autoaugment}.

As we venture into the realm of large language models (LLMs), the significance of data concerns escalates. Research into the scaling laws pertinent to LLMs highlights the critical role of data as a renewable resource crucial for the enhancement and advancement of models \cite{kaplan2020scaling}. With the expansion of model training scales, there is a marked increase in data consumption. Prior studies have examined the escalation in dataset sizes within the domain of machine learning, forecasting that the reservoir of high-quality linguistic data may be depleted by 2026 \cite{villalobos2022will}. This anticipates a potential deceleration in the growth of machine learning models unless there is an enhancement in data efficiency or the discovery of novel data sources. Consequently, the utilization of synthetic data produced by AI models becomes essential once high-quality human-generated data resources are fully exploited. 

From the data perspectives, data augmentation using LLMs offers a viable strategy to overcome these limitations, facilitating the creation of synthetic datasets of high quality that can, in certain instances, exceed the value of data curated by humans \cite{peng2023instruction}. This strategy not only addresses the challenge posed by the limited supply of human-annotated data but also conforms to scaling laws, enabling an increase in the size of training datasets without a proportional escalation in computational costs (FLOPs) \cite{Hoffmann2022TrainingCL}. The tactical employment of synthetic data stands to substantially reduce data collection costs and energy usage, signifying a transformative phase in model training and inference and laying the groundwork for the achievement of artificial general intelligence (AGI) \cite{li2023coannotating}.

From a learning perspective, data augmentation using LLMs has heralded innovative learning paradigms, marking a significant departure from traditional methods predominantly centered around tasks like machine translation, sentiment analysis and NER. The application of LLMs in data augmentation extends to a broader spectrum of learning paradigms, including instruction tuning, in-context learning, and alignment learning. Additionally, it facilitates the generation of pseudo data for classification purposes and the scoring of data for regression analysis. This evolution in methodology not only broadens the scope of potential applications but also invites a deeper exploration into the nuanced benefits and insights that such innovative paradigms offer in the realm of data augmentation. As such, the systematic review of these emerging paradigms is crucial for advancing our understanding and leveraging the full potential of data augmentation in these learning contexts.

\tikzstyle{my-box} = [
    rectangle,
    draw=hidden-black,
    rounded corners,
    text opacity=1,
    minimum height=1.5em,
    minimum width=5em,
    inner sep=2pt,
    align=center,
    fill opacity=.5,
]

\tikzstyle{leaf} = [
    my-box,
    minimum height=1.5em,
    fill=hidden-blue!90,
    text=black,
    align=left,
    font=\normalsize,
    inner xsep=2pt,
    inner ysep=4pt,
]

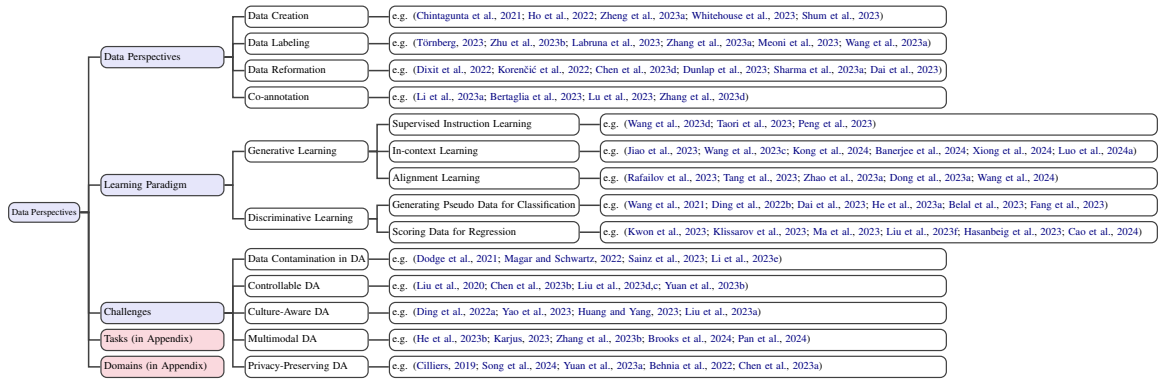
\begin{figure*}
    \centering
    \resizebox{6in}{!}{
        \begin{forest}
            forked edges,
            for tree = {
                grow=east,
                reversed=true,
                anchor=base west,
                parent anchor=east,
                child anchor=west,
                base=left,
                font=\tiny,
                rectangle,
                draw=hidden-black,
                rounded corners,
                align=left,
                minimum width=4em,
                edge+={darkgray, line width=1pt},
                s sep=3pt,
                inner xsep=2pt,
                inner ysep=3pt,
                line width=0.8pt,
                ver/.style={rotate=90, child anchor=north, parent anchor=south, anchor=center},
            },
            where level=1{text width=7.0em, font=\scriptsize}{},
            where level=2{text width=7.0em, font=\scriptsize}{},
            where level=3{text width=11.0em, font=\scriptsize}{},
            where level=4{text width=8.0em, font=\scriptsize}{},
            where level=5{text width=8.0em, font=\scriptsize}{},
            [
                    Data Perspectives,
                    fill=lavender
                    [
                    Data Perspectives,
                    fill=lavender
                    [
                        Data Creation
                        [
                        e.g. \citep{chintagunta2021medically,ho2022large,zheng-etal-2023-augesc,whitehouse2023llm,shum2023automatic}, text width=33em,
                        ]
                    ]
                    [
                        Data Labeling
                        [
                        e.g. \citep{tornberg2023chatgpt,zhu2023chatgpt,labruna2023unraveling,zhang2023prompting,meoni2023large,wang2023t}, text width=33em, 
                        ]
                    ]
                    [
                        Data Reformation
                        [
                        e.g. \citep{dixit-etal-2022-core,korenvcic2022tackling,chen-etal-2023-disco,dunlap2023diversify,sharma-etal-2023-paraphrase,dai2023chataug}, text width=33em, 
                        ]
                    ]
                    [
                        Co-annotation
                        [
                        e.g. \citep{li2023coannotating,bertaglia2023closing,lu2023dialgen,zhang2023toolcoder}, text width=33em, 
                        ]
                    ]
                ]
                [
                    Learning Paradigm,
                    fill=lavender
                    [Generative Learning
                    [
                        Supervised Instruction Learning
                        [
                        e.g. \citep{wang-etal-2023-self-instruct,alpaca,peng2023instruction}, text width=33em,
                        ]
                    ]
                    [
                        In-context Learning
                        [
                        e.g. \citep{jiao2023panda,wang2023augmenting,kong2024audio,banerjee2024context,xiong2024large,luo2024chain}, text width=33em, 
                        ]
                    ]
                    [
                        Alignment Learning
                        [
                        e.g. \citep{rafailov2023direct,tang2023just,zhao2023group,dong2023aligning,wang2024weaver}, text width=33em, 
                        ]
                    ]
                    ]
                    [Discriminative Learning
                    [
                        Generating Pseudo Data for Classification
                        [
                        e.g. \citep{Wang2021WantTR,ding2022gpt,dai2023chataug,he2023teacherlm,belal2023leveraging,fang2023chatgpt}, text width=33em, 
                        ]
                    ]
                    [
                        Scoring Data for Regression
                        [
                        e.g. \citep{kwon2023reward,klissarov2023motif,ma2023eureka,liu2023learning,hasanbeig2023allure,cao2024drlc}, text width=33em, 
                        ]
                    ]
                ]]
                [
                    Challenges,
                    fill=lavender
                    [
                        Data Contamination in DA
                        [
                        e.g. \citep{dodge2021documenting,magar2022data,Sainz2023NLPEI,Li2023LatestEvalAD}, text width=33em,
                        ]
                    ]
                    [
                        Controllable DA
                        [
                        e.g. \citep{Liu2020TellMH,Chen2023MixtureOS,Liu2023BenchmarkingGA,liu2023context,Yuan2023LLMFP}, text width=33em, 
                        ]
                    ]
                    [
                        Culture-Aware DA
                        [
                        e.g. \citep{ding-etal-2022-globalwoz,yao2023empowering,huang-yang-2023-culturally,liu2023multilingual}, text width=33em, 
                        ]
                    ]
                    [
                        Multimodal DA
                        [
                        e.g. \citep{He2023UsingAS,Karjus2023MachineassistedMM,zhang2023video,videoworldsimulators2024,pan2024unifying}, text width=33em, 
                        ]
                    ]
                    [
                        Privacy-Preserving DA
                        [
                        e.g. \citep{Cilliers2019WearableDI,song2024llm,yuan2023large,behnia2022ew,chen2023federated}, text width=33em, 
                        ]
                    ]
                ]
                [
                Tasks (in Appendix),
                fill=palepink
                ]
                [
                Domains (in Appendix),
                fill=palepink
                ]
            ]
        \end{forest}
    }
    \caption{Taxonomy of Data Augmentation using LLMs. Tasks and Domains are included in the Appendix.}
    \label{fig:taxonomy}
\end{figure*}

Given the growing interest and work in this domain, we believe it is a timely moment to present a paper on LLM-based DA. This paper aims to: (i) discuss data augmentation using LLM from the data perspective (ii) explore the learning paradigms that involve training LLMs on data generated by LLMs themselves, and (iii) highlight the principal challenges in this field to effectively guide and spur further interest and research. To the best of our knowledge, this represents the first survey that delves into data augmentation methods leveraging LLMs in such detail, thereby marking a significant contribution to the literature on LLM applications.

This paper is structured as follows. Section~\ref{sec:related_works} provides a comprehensive review of related surveys, highlighting the distinctions between this work and previous surveys. Following that, as illustrated in Figure~\ref{fig:taxonomy}, Section~\ref{sec:methods} presents analysis from data perspectives for data augmentation with LLMs, establishing a foundational understanding for the community. Section~\ref{sec:learning} explores the learning paradigms associated with data augmentation using LLMs, neatly categorizing existing methods into two primary types: generative learning and discriminative learning. Section~\ref{sec:challenges} ventures into the challenges and prospective future directions for research in this domain, highlighting the complexities and opportunities that lie ahead. To further aid the community, the Appendix offers a detailed listing of existing methods categorized by tasks (Section~\ref{sec:tasks}) and domains (Section~\ref{sec:domains}), serving as a valuable resource for researchers and practitioners alike. Through this survey, we hope to attract wider attention, generate increased interest, and encourage further research in the field of data augmentation using LLMs.

\section{Related Surveys}
\label{sec:related_works}
% \textbf{Bold form + Points: 1. Pre-LLM vs LLM Data augmentation 2. Instruction-tuning}

\paragraph{Pre-LLM DA vs LLM DA}
Prior to the advent of Large Language Models (LLMs), surveys like those by \citet{Feng2021ASO} and \citet{hedderich-etal-2021-survey} explore data augmentation in NLP, highlighting methodologies, applications, and the challenges of low-resource NLP. However, the surveyed approaches are constrained by the need for a deep linguistic understanding of traditional NLP tasks and do not fully utilize LLMs. The emergence of LLMs has revolutionized this landscape by significantly enhancing the quality of synthetic data, leading to innovative methodologies and broader applications. LLMs' nuanced understanding of language has mitigated previous limitations such as the issue of synthetic data being of poor quality, containing significant noise, and detrimentally affecting model performance. This advancement has made data augmentation more effective and accessible. \citet{chen-etal-2023-empirical} provide an empirical analysis of various augmentation techniques in supervised and semi-supervised contexts, focusing on pre-LLM methods. However, unlike previous surveys, we focus on discussing \textit{data augmentation using LLMs from the data and learning perspectives}.

% There have been previous surveys covering data augmentation techniques in NLP before the era of LLMs. \citet{feng-etal-2021-survey} surveys data augmentation techniques in NLP. It discusses data augmentation in terms of methodology, applications, and tasks. As the data quality generated by LLMs has significantly improved, new methodologies and applications have been proposed.
% \citet{hedderich-etal-2021-survey} discusses the different dimensions of data availability and surveys promising approaches for low-resource NLP, including data augmentation techniques. However, they conclude that, as data augmentation requires an in-depth understanding of the language and task, it has not been used widely within the NLP community. This issue has been alleviated by the introduction of LLMs. \cite{chen2023empirical} focus on the application of data  to learning from limited
% data by providing an empirical study over different augmentation methods on various benchmark
% datasets in both supervised and semi-supervised
% settings. However, the survey only focused on the methods before LLMs.

% While the methodology has shifted with the development of LLMs, the applications remain similar, including low-resource languages, mitigating bias, and few-short learning.

\paragraph{Instruction-tuning \& Alignment} 
Recent advancements in LLMs have heralded the advent of novel learning paradigms, including instruction-tuning and alignment learning, to better align the objectives of model training with the user's expectations for adherence to instructions. Instruction tuning (IT) emerges as a potent strategy to enhance the efficacy and manageability of LLMs, as discussed in the survey by \citet{zhang2023instruction}, which identifies and debates the hurdles encountered by IT, such as generating high-quality instruction sets, the method's effectiveness on tasks not initially supported, and critiques concerning its tendency for shallow pattern recognition. This underscores an imperative for ongoing research and refinement in the domain of instruction fine-tuning. Meanwhile, \citet{ji2023ai} delineate the foundational principles, methodologies, and practical applications of AI alignment, exploring its prospective trajectories. The focal point of AI alignment research is to ensure AI systems' actions are congruent with human objectives and ethical standards, tackling misalignment through enhancements in system robustness, interpretability, controllability, and ethical considerations, despite obstacles like reward exploitation and goal misinterpretation. Our survey concentrates on this emerging learning paradigm, particularly the use of synthetic data to train LLMs and augment their performance capabilities.

\section{Data Perspectives}
\label{sec:methods}
% Ruochen

From a data perspective, we group existing studies on LLM-based DA into four categories: 1. \textit{Data creation} which leverages the few-shot learning ability of LLMs to create a large synthetic dataset; 2. \textit{Data labeling} which uses the LLM to label existing datasets; 3. \textit{Data reformation} which transforms existing data to produce new data; 4. \textit{Co-Annotate} which enables LLM-human collaboration to gather high-quality augmentation data. This section discusses relevant papers in each category.%In this section, we discuss relevant papers for each method.

\subsection{Data Creation}

Data Creation focuses on leveraging the few-shot learning ability of LLMs to quickly create a large amount of synthetic data. It is most used in tasks with a large label space.

Data Creation with LLMs is a promising solution in specialized or private domains, where annotations are usually difficult or expensive to collect. Dialogue tasks are one example where specialized data is hard to collect. In medical dialogue summarization, \citet{chintagunta2021medically} uses a powerful few-shot learner such as GPT-3 to create synthetic medical dialogue summaries. By training models on a mix of synthesized and human-labeled data, the algorithm can scale a few human-labeled examples to yield results comparable to using 30x human-labeled examples. {Similarly,} for general dialogue, Dialogic~\citep{li-etal-2022-controllable} is seeded with a few dialogues and can automatically select in-context examples for demonstration and prompt LLMs to generate annotated dialogues in a controllable way. Then, automatic verification and revision methods are proposed to mitigate annotation errors. Results show that performance greatly improves in low-resource scenarios. \citet{wan-etal-2022-unified} also attempt few-shot data augmentation on dialogue modeling. {Aside from few-shot learning, }for emotional support conversations, AugESC~\citep{zheng-etal-2023-augesc} finetunes an LM and prompts it to complete dialogues from collected posts. The post-training on AugESC improves downstream dialogue models' generalization abilities to open-domain topics. 
For low-resource classification, {LLM can be used to create synthetic examples of a given label.}
% To improve few-shot classification tasks, % already mentioned 'low-resource classification'
\citet{moller2023prompt} gives an example and its corresponding label and instruct the LLMs to generate similar examples exhibiting the same label. Resulting models yield better downstream performances {on few-shot classification} but still lag behind human-annotated data. 
{For other low-resource tasks such as recommendation and intent detection, Data Creation can also effectively boost the training data space.}
To gather better recommendations, \citet{zhang2023recommendation} generates a large amount of user-personalized instruction data with varying preference and intention types. Then, the LLM is optimized using instruction tuning. The resulting model can obtain more accurate recommendations and outperform competitive baselines, including GPT-3.5. 
For intent detection, \citet{lin-etal-2023-selective} first uses an LLM to generate synthetic examples in the context of the training set and then uses Pointwise V-Information (PVI) to filter unhelpful examples. \citet{sahu-etal-2022-data} prompts GPT-3 to generate labeled training data, which can significantly boost the intent classifier's performance for distinct intents but becomes less helpful with semantically close intents.

Data Creation also helps in more general tasks {by generating new training datasets}. For information retrieval, \citet{bonifacio2022inpars} uses few-shot prompting with LLMs to generate synthetic training datasets consisting of query-document pairs. Retrieval models finetuned with the augmented data significantly outperform unsupervised models. For reasoning, Logi-CoT~\citep{liu2023logicot} gathers a new instruction-tuning dataset by prompting GPT-4 and is used for teaching models to elicit general reasoning skills.

{Moreover, Data Creation is helpful for model performance distillation.} To distill LLM{s' reasoning} performances to smaller models, Fine-tune-CoT~\citep{ho2022large} uses zero-shot CoT prompting to generate rationales from teacher models and use them to fine-tune smaller student models. The resulting performance improvements are stable across dataset size, teacher performance, etc. To reduce the need of manual annotation in reasoning tasks, Automate-CoT~\citep{shum2023automatic} automatically generates pseudo-CoTs from a small labeled dataset and then prunes and selects an optimal combination for CoT prompting. Similarly {for instruction-following}, \citet{peng2023instruction} uses GPT-4 to generate an instruction-following dataset and feedback data. The resulting instruction-tuned LLaMa models can lead to comparable performance with the original GPT-4. To aid multilingual commonsense reasoning tasks, \citet{whitehouse2023llm} provides LLMs with instructions and examples from the original training data, prompting them to generate new and diverse examples. By training with augmented data, significant cross-lingual performance improvements are observed on smaller models.

To systematically study the behavior of such data creation methods and improve upon current few-shot prompting methods, \citet{meng2023tuning} attempts to first tune an LM on few-shot examples and then use it as a generator to synthesize a large amount of novel training samples. The resulting approach could augment task performances than existing few-shot learning methods.

\subsection{Data Labeling}

Data Labeling seeks to utilize the general language comprehension abilities of LLMs to annotate unlabeled datasets. It is primarily useful in tasks that have a large enough unlabeled data corpus, such as cross-lingual and multimodal tasks.

To evaluate LLMs' potential in data labeling, \citet{tornberg2023chatgpt} studies the zero-shot annotation ability of GPT-4 on labeling political twitter messages with political tendancies. Compared to human workers, the LLM annotations display higher accuracy and lower bias. Similarly, \citet{zhu2023chatgpt} observes that, in social computing tasks, ChatGPT has the potential to accurately reproduce human labels. Notably, annotations from open-source LLMs~\citep{alizadeh2023open} and ChatGPT~\citep{Gilardi_2023} can surpass crowd-worker performance on annotation tasks. For annotating low-resource tasks such as goal-oriented dialogues~\citep{labruna2023unraveling} and speech emotional data~\citep{latif2023can}, the quality of ChatGPT annotations is on par with human-generated labels. However, \citet{bansal2023large} observes that simply annotating can sometimes worsen generalization. Thus, it proposes conditional sampling to optimize the tradeoff between informativeness and budget

Cross-lingual {tasks mostly contain a large unlabeled corpus, which could benefit from data labeling. Therefore,} \citet{zhang2023prompting} uses different prompting strategies {to augment} machine translation (MT) {data}. It tries augmenting monolingual data using back-/forward-translation via zero-shot prompting, which still suffers from limitations such as generalization and unstable transfer performances. {Similarly, }\citet{meoni2023large} annotates training data for multilingual clinical entity extraction with LLMs. After fine-tuning smaller models with augmentations, they display promising results for information extraction (IE) tasks. 

Data labeling is also promising for multimodal applications. For data-scarce Visual Question Answering (VQA) tasks, \citet{khan2023q} utilizes a Self-taught Data Augmentation (SelTDA) framework to generate pseudo labels from unlabeled images. The pseudo-labeled data could then improve VQA task performance and robustness. Combining multi-modality with reasoning, T-SciQ~\citep{wang2023t} further distills LLMs' reasoning abilities in multimodal tasks by asking the teacher model to produce CoT rationales. As a result, it achieves state-of-the-art performance in scientific QA.

\subsection{Data Reformation}

Data Reformation techniques attempt to reformulate the existing data into more variations for more fine-grained augmentation. 

Such reformation techniques could naturally aid in counterfactual generation tasks{, which reforms existing data to its counterfactual version}. Disco~\citep{chen-etal-2023-disco} uses LLMs to generate high-quality counterfactual data at scale. It first uses in-context learning with GPT-3 to generate phrasal perturbations, then uses a task-specific teacher model to filter and distill high-quality counterfactual data. Models trained using the generated counterfactuals display improved robustness and generalization across distributions. For retrieval-augmented generation, CORE~\citep{dixit-etal-2022-core} uses GPT-3 to generate counterfactual edits to the input conditioned on the retrieved excerpts. The perturbations then help mitigate model bias and improve performance on out-of-distribution (OOD) data. 

Data Reformation could also quickly diversify the original dataset {by forming data pairs}. For conspiracy detection, \citet{korenvcic2022tackling} asks GPT-3 to rephrase tweets with original labels to augment training. To generate useful variations of the pretraining datasets of large vision models, ALIA~\citep{dunlap2023diversify} uses LLMs to generate image descriptions and augment the training data via language-guided image editing. By leveraging LLMs to the image domain, ALIA surpasses traditional data augmentation methods on fine-grained classification tasks. For Named Entity Recognition (NER), \citet{sharma-etal-2023-paraphrase} generates paraphrases while retaining inline annotation for entities. Among other PLMs, GPT-3 is able to generate high-quality paraphrases, yielding statistically significant improvements in NER performance.

For more general tasks, {Data Reformation could help to diversify and broaden the original dataset.} AugGPT~\citep{dai2023chataug} tries to overcome the challenge of few-shot and data-scarce NLP tasks by rephrasing each sentence in the training samples into 6 semantically similar sentences. Experiments show that such an approach surpasses state-of-the-art text data augmentation methods in augmentation distribution and testing accuracy. For effective knowledge distilation, GPT3Mix~\citep{yoo-etal-2021-gpt3mix-leveraging} extracts sample sentences from the task-specific training data, embed these samples in the prompt, and asks the LLM to generate an augmented mixed sentence influenced by the sample sentences. \citet{guo2023dr} asks GPT-3.5 and GPT-4 to rewrite or generate question-answer pairs with zero-shot prompting. Fine-tuning with the refined and diversified training set then successfully distils medical question-answering abilities to smaller models. 

\subsection{Co-annotation}

Co-annotation refers to the collaborative annotation process between humans and LLMs. By combining both annotation approaches, Co-annotation can reduce annotation costs and improve annotation performance at the same time. {Firstly, }\citet{li2023coannotating} proposes CoAnnotating, which allocates a given datapoint to be annotated by humans or by LLMs by computing the uncertainty level of LLM's annotations. With efficient human-AI collaboration, it provides insights into the tradeoff between annotation quality and annotation cost. {To assist human annotators with explanations, }\citet{bertaglia2023closing} asks the LLM to identify relevant features, such as text tokens, as assistive explanation. The approach improves inter-annotator agreement, annotation accuracy, and annotators' confidence, eventually leading to more transparency. {Using human feedback to direct LLM annotations could also effectively generate high-quality data. }Diagen~\citep{lu2023dialgen} uses an LLM to iteratively generate dialogues in protected data domains, where human feedback is used to correct inconsistencies or redirect the flow in sub-dialogues. As a result, fine-tuning or in-context learning with the annotated data shows significant model performance improvements. {Similarly, }ToolCoder~\citep{zhang2023toolcoder} uses human-written input-output pairs as prompts to guide chatGPT to annotate a tool-augmentation dataset. Then, the annotated data is filtered to ensure quality. After fine-tuning with the annotated data, ToolCoder can achieve comparable performance with ChatGPT on code generation. 

% \subsection{Comparisons}
% \jh{Seems not very connected to the previous subsections.}
% \citet{ding-etal-2023-gpt} investigates the ability of GPT-3 to serve as a data annotater. Specifically, GPT3 is prompted to perform 3 annotation tasks: prompt-guided unlabeled data annotation (PGDA); prompt-guided training data generation (PGDG); and dictionary-assisted training data generation (DADG). Results show that annotation is suitable for tasks with small label space while generation-based methods are more suitable for tasks with large label space. Specifically, generation-based approaches tend to be more cost-effective. Similarly, many other works~\citep{ubani2023zeroshotdataaug} attempt to evaluate LLM's data augmentation performances.
\section{Learning Paradigms}
\begin{figure}[t!]
    \centering
    \includegraphics[scale=0.4]{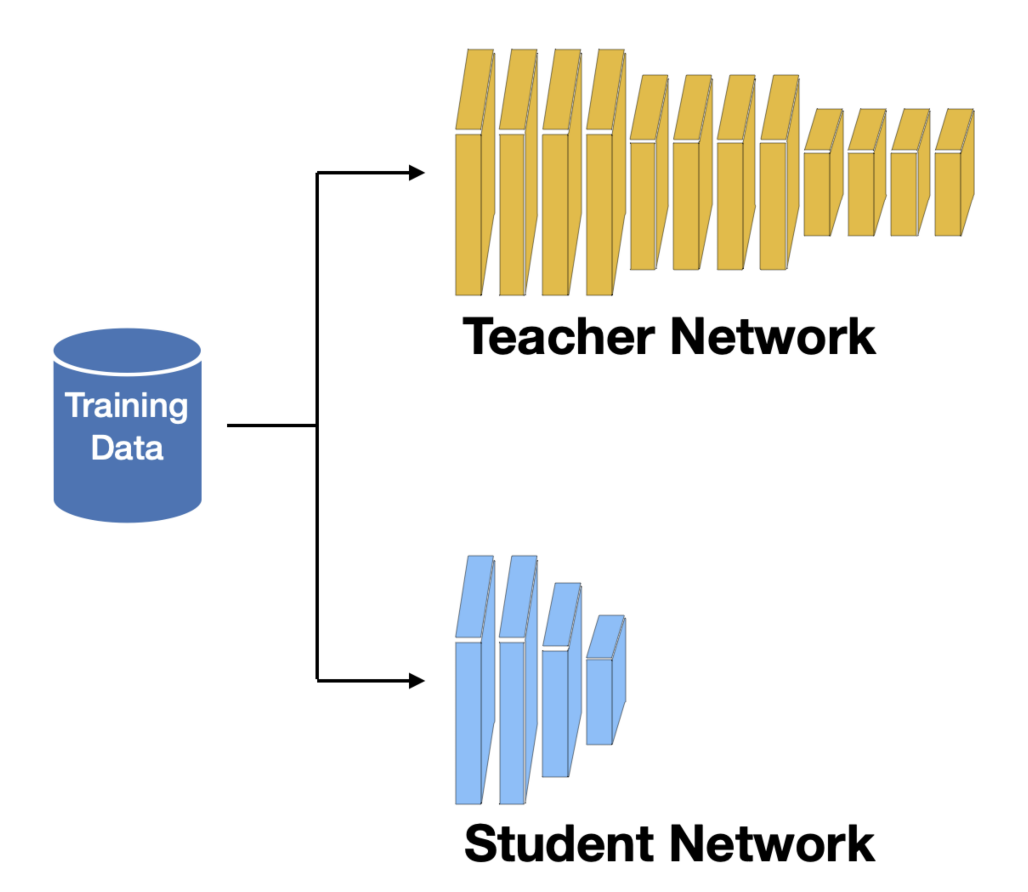}
    \caption{Illustrations of Teacher-Student Learning.}
    \label{fig:method} 
\end{figure}

% \textbf{Change the point list to paragraph}
\label{sec:learning}
\paragraph{Teacher-Student Learning (TSL)}
Leveraging a Language Model (LLM) as a data annotator to create synthetic data for the purpose of training other models represents a pivotal shift towards more efficient and scalable machine learning methodologies. This approach is situated within the broader framework of \emph{Teacher-Student Learning (TSL)} \cite{hu2022teacher}, where the LLM acts as the 'teacher' by generating annotated data that serves as training material for 'student' models as shown in Figure~\ref{fig:method}. By utilizing an LLM for data annotation, organizations can bypass some of the limitations associated with human annotators, such as scalability issues and subjectivity, thereby streamlining the development process of AI systems. This innovative strategy not only enhances the efficiency of generating training datasets but also contributes to the advancement of machine learning techniques, enabling the creation of more robust and accurate models. This emerging trend is gaining traction, particularly in the context of training large language models.
% Use LLM as a data annotator (instead of a human annotator) to generate synthetic data for training other models. Essentially, this falls under the big umbrella of the \emph{Teacher-Student Learning} (TSL) paradigm.

% \subsection{Teacher-Student Learning (TSL)}
 Teacher-Student Learning paradigms for data augmentation using LLMs can be broadly segmented into generative and discriminative learning approaches. Generative learning exploits LLMs to create instructional datasets and demonstration examples, enriching model training. It encompasses Supervised Instruction Learning, generating instructional data \((\text{instruction}, x, y)\); In-context Learning, producing few-shot examples to augment query inputs and facilitate output generation; and Alignment Learning, creating human-preference data for reinforcement learning and optimization. Discriminative learning, conversely, focuses on refining task-specific models through pseudo-labeling and pseudo-scoring. Pseudo-labeling generates labels \((x, \hat{y})\) for smaller models, akin to knowledge distillation, while pseudo-scoring assigns numerical scores to \((x, y)\) pairs for regression tasks.

\subsection{Generative Learning}
% \begin{itemize}
%     \item 
    
\paragraph{Supervised Instruction Learning}: Supervised Instruction Learning, leveraging LLMs for data augmentation, presents a novel approach to enhancing instructional data generation. The generation process entails creating instructional data in the form of (instruction, x, y) triplets, where 'x' represents optional context or input, and 'y' denotes the output generated by LLMs in response to the given instruction. \citet{wang-etal-2023-self-instruct} 
 first introduced "Self-Instruct," a method to enhance the instruction-following ability of pre-existing language models by leveraging their own outputs. \citet{alpaca} created the Alpaca dataset at low cost using Supervised Instruction Learning and \citet{peng2023instruction} further explored the potential of leveraging LLMs to create cross-lingual instruction following data, self-evaluation and cross-model assessment data and responses to unnatural instructions.
    
    % Generate instructional data, for example, (instruction, $x$, $y$).

\paragraph{In-context Learning}: In-context learning represents a new paradigm where models, particularly LLMs like GPT-3, generate predictions based on examples provided within the context of the input, without requiring explicit retraining or parameter updates \cite{liu2022few, Dong2023ASF,zhang2024makes}. To enable LLMs with such a capability, data with rich context should be provided to LLMs during the training process.  \citet{kong2024audio} use existing LLMs to generate two multiturn dialogue datasets based on detailed annotations, enabling model's in-context dialogue capabilities. \citet{banerjee2024context} utilize KGs to build up in-context training samples for LLM instruction tuning. \citet{xiong2024large} construct a synthetic temporal QA dataset to enhance the LLM's capability in temporal QA tasks. \citet{luo2024chain} design structure-based augmentation methods with LLMs for Temporal Knowledge Graph Completion. It is a common trend that various models adopt LLMs to generate synthetic data with long context to enhance the model's in-context dialogue, inference, QA, and Natural Language Generation (NLG) capabilities \cite{touvron2023llama,jiao2023panda,wang2023augmenting}.

    % involves dynamically updating the model's parameters during inference based on the context of the input, allowing it to adapt and generate responses that better match the current conversation or task at hand.
    
    % Generate few-shot demonstration examples, for example, $(x_1, y_1, x_2, y_2, \ldots, x_n, y_n)$ to augment a query input $x_q$, and prompt the model to generate $y_q$. This checks the zero-shot Chain of Thought (CoT), adopting a step-by-step thought process. 

    % \item 
\paragraph{Alignment Learning}: Alignment learning focuses on training the model to align its outputs with human preferences or objectives, often through techniques such as reward shaping or reinforcement learning, to optimize its performance towards desired outcomes while minimizing unintended behaviors. \citet{rafailov2023direct} propose the Direct Preference Optimization (DPO) method to effectively train LLMs on alignment learning. \citet{wang2024weaver} adopt DPO and synthesized instruction following data to build a foundation model for creative writing. \citet{tang2023just} utilize LLMs such as ChatGPT, GPT-4, Dolly-v2, and StableVicuna to generate various data for alignment training of their model LLMDA. \citet{zhao2023group} introduce Group Preference Optimization (GPO) to the alignment training for LLMs. \citet{dong2023aligning} adopt data augmentations from LLMs and construct contrastive feedback to training models.

    % Can we generate human-preference data for Reinforcement Learning from Human Feedback (RLHF), Deep Preference Optimization (DPO)?  
    
    % For example, $(x, y_1 > y_2 > y_3 \ldots)$. 
% \end{itemize}

\subsection{Discriminative Learning}
% \begin{itemize}
    %\item \textbf{Pseudo-labeling for Classification} 
    %(Traditional knowledge distillation): Generate pseudo labels $(x, \hat{y})$ for training a smaller discriminative model (like a BERT-style model). investigate
\paragraph{Generating Pseudo Data for Classification} 
    LLMs play an important role in generating pseudo data to train discriminative models for classification tasks. Many LLM-based data augmentation approaches fall into this category. \citet{Wang2021WantTR,ding2022gpt} explore using GPT-3 to annotate or generate data to train smaller task-specific models on a variety of classification tasks. \citet{dai2023chataug} propose AugGPT which uses ChatGPT to rephrase training samples, surpassing state-of-the-art text data augmentation methods on few-shot classification tasks. To enhance the capabilities of student models in zero-shot learning, \citet{he2023teacherlm} develop TeacherLM7.1B which demonstrates the strong ability to generate augmentations for training student models. Besides, some works are designed for specific target tasks, \eg\ sentiment analysis \citep{belal2023leveraging,moller2023prompt} and intent classification \citep{sahu2022data,fang2023chatgpt}.

    %\item \textbf{Pseudo-scoring for Regression} 
    %(Reward modeling, evaluation scoring): Generate a numerical score for a pair of $(x, y)$.
 \paragraph{Scoring Data for Regression} LLMs can serve as an essential solution for scoring the input. \citet{kwon2023reward} explore prompting GPT-3 as a proxy reward function in a reinforcement learning framework. The reward signal is then used to update the behavior of the agent. Motif \citep{klissarov2023motif} proposes to elicit preferences from an LLM over paired data to construct rewards for training agents. \citet{ma2023eureka} introduce Eureka which can leverage GPT-4 for evolutionary optimization to generate reward functions that outperform expert-designed rewards. To mitigate reward sparsity that can lead to inefficient and unstable learning, \citet{cao2024drlc} incorporate a critic LLM to provide more fine-grained dense rewards. In addition to generating reward signals, LLMs are also used to evaluate different properties of text. \citet{liu2023learning} investigate utilizing LLMs as the reference to evaluate the quality of the generated summary. \citet{liu2023benchmarking} further benchmark LLM-based evaluation for instruction controllable text summarization with different evaluation protocols and LLMs. To improve LLMs' text evaluation capabilities, ALLURE \citep{hasanbeig2023allure} explores leveraging iterative in-context learning to audit and enhance LLM-based evaluation. 
\section{Challenges and Future Directions in Data Augmentation using LLMs}
\label{sec:challenges}
% \textbf{Change to Challenges vs Future Directions}
\subsection{Data Contamination in Data Augmentation}
\paragraph{Challenges} In leveraging LLMs for data augmentation, the risk of data contamination presents a significant challenge \cite{magar2022data}. This issue, where training data inadvertently includes evaluation set examples, undermines model evaluation integrity by enabling models to memorize rather than generalize. The two main types of contamination are input contamination and the more severe input-and-label contamination, which allows models to memorize exact input-output pairs. Studies such as \cite{dodge2021documenting} and others \cite{Sainz2023NLPEI, Li2023LatestEvalAD} highlight the prevalence of data contamination across various NLP benchmarks, emphasizing the need for effective detection and mitigation strategies. The integration of LLMs for data augmentation, while offering the potential to enrich training datasets and enhance model performance, necessitates cautious consideration of the risk of data contamination. This risk not only undermines the integrity of model evaluations but also highlights the ongoing need for novel strategies to detect and counteract data contamination. 

 \paragraph{Future Directions} As the NLP field continues to advance, addressing the challenges posed by data contamination, particularly in the context of LLM-augmented datasets, will be crucial for maintaining the credibility and effectiveness of machine learning models. To this end, there is a significant space for research on the development of innovative methodologies and tools specifically designed to detect and address data contamination in synthetic datasets.

\subsection{Controllable Data Augmentation}
\paragraph{Challenges} Controllable data augmentation using LLMs faces significant challenges, primarily due to difficulties in ensuring the quality of generated synthetic data \cite{Liu2020TellMH, Chen2023MixtureOS}. These techniques, aimed at enriching few-shot or low-resource datasets by targeting specific attributes, rely on models understanding and applying attribute-value mappings. Current methods, including in-context learning and fine-tuning, often struggle with maintaining quality across non-target dimensions during augmentation \cite{Liu2023BenchmarkingGA}. Furthermore, model collapse is a critical challenge that arises when models learn from data generated by other models, leading to a degeneration where models progressively lose information about the true underlying data distribution \cite{shumailov2023curse}. Over time, this process results in models that forget the true underlying data distribution and converge to a narrow interpretation of the data with very little variance, potentially compromising the diversity and richness of the model's output. 

\paragraph{Future Directions} 
 To address these challenges, we need to explore new methods that can decompose and reconstruct texts with LLMs to precisely control attribute changes. To mitigate the issues of model collapse, the generated data should follow the data distribution of human-generated data to preserve integrity and reliability. Moreover, the generated synthetic data must be approached with a keen awareness of the potential for bias and the ethical implications associated with LLM outputs. By adopting responsible augmentation strategies that promote diversity and implement ethical guidelines, such as integrating the in-context vector \cite{liu2023context} into the data augmentation framework, we can mitigate biases and advance NLP in a manner that is technologically effective and ethically responsible \cite{liu2023context,Yuan2023LLMFP}.

% To address these challenges, we need to explore new methods that can decompose and reconstruct texts with LLMs to precisely control attribute changes. Moreover, the generated synthetic data must be approached with a keen awareness of the potential for bias and the ethical implications associated with LLM outputs. By adopting responsible augmentation strategies that promote diversity and implement ethical guidelines, such as integrating the in-context vector into data augmentation framework \cite{liu2023context}, we can mitigate biases and advance NLP in a manner that is technologically effective and ethically responsible . 

% In the realm of future directions in data augmentation using Large Language Models (LLMs), a formidable challenge that persists is the development of controllable data augmentation techniques. While LLMs have demonstrated remarkable capabilities in generating augmented data, a fundamental issue arises from the inherent dissimilarity between AI-generated data and human-written data, making it challenging for human observers to discern the differences. To advance in this field, it is imperative to devise methodologies that can quantize the validity and diversity of augmented data within the distribution. This not only ensures the reliability of the augmented data but also paves the way for more effective and robust data augmentation strategies, addressing the ongoing need for enhancing the performance and generalization of machine learning models in various domains.

\subsection{Culture-Aware Multilingual Data Augmentation}
\paragraph{Challenges} In the rapidly evolving domain of multilingual NLP systems, the integration of culture-aware data augmentation emerges as a pivotal challenge \cite{yao2023empowering,huang-yang-2023-culturally}. This approach is critical for the effective localization of these systems, enabling them to operate seamlessly across varied real-world contexts \cite{ding-etal-2022-globalwoz,liu2023multilingual}. Traditional data augmentation methods, which predominantly focus on linguistic variations, often fail to capture the complex cultural nuances that significantly influence language usage. This oversight highlights the need for a fundamental shift towards embedding cultural intelligence into data augmentation strategies. Utilizing LLMs for this purpose offers a promising pathway.

\paragraph{Future Directions} LLMs have the potential to generate data that reflects cultural specifics, encompassing regional idioms, social norms, and linguistic nuances. Such culturally enriched data augmentation not only aims to improve the precision of multilingual NLP systems but also promotes inclusivity and global relevance. Nevertheless, achieving this necessitates further research in developing LLMs capable of discerning and adjusting to cultural differences. This advancement is crucial for realizing NLP solutions that are not only linguistically diverse but also culturally attuned, thereby enhancing their acceptability and usability on a global scale. The journey towards incorporating cultural awareness in data augmentation with LLMs presents both a formidable challenge and a significant opportunity for the future of multilingual NLP systems.

\subsection{Multimodal Data Augmentation}
\paragraph{Challenges} In the burgeoning field of multimodal data augmentation using LLMs, researchers face a constellation of challenges despite the promising prospects \cite{He2023UsingAS,Karjus2023MachineassistedMM}. As LLMs evolve to incorporate a diverse array of data types beyond text—spanning images, audio, video, and potentially graphs—the complexity of integrating and harmonizing these varied modalities poses significant technical hurdles \cite{zhang2023video,videoworldsimulators2024}. The seamless handling and enhancement of the intricate interplay between these modalities require sophisticated algorithms that can not only generate and manipulate multimodal data but also understand and preserve the contextual and semantic integrity across them. Additionally, the extension to graph-based modalities introduces unique challenges in representing and leveraging relational and structural information effectively \cite{pan2024unifying}. 

\paragraph{Future Directions} The challenges mentioned above demand innovative approaches in model architecture, data representation, and algorithmic efficiency to fully realize the potential of LLMs in multimodal augmentation. Moreover, ensuring the adaptability and robustness of these models in real-world applications across domains such as social network analysis and recommendation systems amplifies the complexity of the task. As we push the boundaries of what LLMs can achieve in multimodal learning, addressing these challenges will be crucial for unlocking new horizons in research and innovation, driving forward the capabilities of machine learning models to comprehend and process the rich tapestry of multimodal information.

\subsection{Privacy issues of Data Augmentation}

\paragraph{Challenges}  Data augmentation using LLMs presents significant privacy concerns, especially in sensitive sectors like healthcare \cite{Cilliers2019WearableDI}, where confidentiality is paramount. While LLMs offer the potential to enrich sparse datasets with synthetic data, safeguarding sensitive information within these datasets poses considerable challenges. Research by \citet{song2024llm} underscores the limitations of Differential privacy (DP) in effectively generating synthetic data without compromising privacy or data quality. Concurrently, \citet{yuan2023large} explore LLMs' utility in healthcare for patient-trial matching, proposing a privacy-aware approach that mitigates privacy risks by using desensitized data. These studies highlight the ongoing struggle to leverage LLMs for data augmentation while ensuring robust privacy protection, indicating a crucial area for further research and development in privacy-preserving techniques.

\paragraph{Future Directions}
Future directions to mitigate these privacy issues may involve developing more robust anonymization techniques to ensure that no traceable data is reproduced in the outputs. Additionally, implementing differential privacy \cite{behnia2022ew} and federated learning \cite{chen2023federated} could help minimize privacy risks by allowing data to be processed in a way that doesn't reveal identifiable information. Transparent data usage policies and regular audits could also play a crucial role in maintaining user trust and ensuring compliance with evolving data protection regulations. These measures would be essential in balancing the benefits of LLMs in data augmentation with the need to protect individual privacy.

\section{Conclusion}
In this survey, we have delivered an in-depth and organized review of data augmentation techniques utilizing Large Language Models (LLMs). We explored key data augmentation strategies and the application of LLMs in enhancing data augmentation processes. Moreover, we discussed the learning paradigm under the generative and discriminative learning perspectives. Additionally, we highlighted the existing challenges and potential future research opportunities, indicating significant potential for further developments in this field. Our aim is for this paper to act as a reference for AI researchers, aiding them in selecting appropriate data augmentation methods and encouraging further investigation and enthusiasm in this domain.

\section{Limitations}
Data augmentation using Large Language Models (LLMs) presents several limitations that can impact their effectiveness in certain contexts. Firstly, the quality of the generated data heavily depends on the training data and the model's architecture, which can lead to biases or inaccuracies being introduced into the augmented dataset. This can perpetuate or even exacerbate existing biases in the training data, leading to skewed or unfair outcomes in applications. Secondly, LLMs may struggle with generating high-quality data for highly specialized or niche domains where specific knowledge or terminology is required, due to the limited exposure of such content in their training data. Additionally, the cost of generating large amounts of augmented data can be prohibitive, as it requires substantial computational resources, especially for state-of-the-art models. Lastly, there's the challenge of ensuring the semantic consistency and uniqueness of the generated data, as LLMs can sometimes produce repetitive or generic outputs that may not add meaningful diversity to the dataset. These limitations necessitate careful consideration and mitigation strategies to effectively leverage LLMs for data augmentation purposes.

\bibliography{anthology_p2, acl2023}
% \bibliography{anthology, acl2023}
\bibliographystyle{acl_natbib}

\appendix
\section{Tasks} \label{sec:tasks}
In this section, we classify the works based on the task that data augmentation is used to solve. We categorize the tasks as fundamental tasks, understanding based tasks, generation tasks, multi-modal tasks, and other tasks. We elaborate on the fundamental tasks in \cref{sub:basic tasks}, which includes text classification, machine translation, and sequence tagging. We explain the understanding and inference tasks in \cref{sub:understanding}, with question answering, logical reasoning, and natural language inference. We discuss the generation and interaction tasks in \cref{sub:generation}, which includes summarization, data-to-text NLG, open-ended and conditional generation, and dialog. We introduce multi-modal tasks in \cref{sub:multimodal}, which includes story telling and VQA tasks. Lastly, we summarize some other tasks that do not belong to above mentioned categories in \cref{sub:others}. 

\subsection{Fundamental Tasks} \label{sub:basic tasks}
In this subsection, we discuss some fundamental tasks in NLP. These tasks are considered relatively simple to the powerful LLMs nowadays. LLMs are able to achieve human-level performance in tasks like machine translation. \citet{ding2022gpt} studies 3 different approaches to utilize GPT-3 for data augmentation: 1) prompt-guided unlabeled data annotation (PGDA); 2) prompt-guided training data generation (PGDG); and 3) dictionary assisted training data generation (DADG). Results show that it is largely reliable to count on the LLMs as data augmentation tools. 

\paragraph{Text Classification}
Text classification involves categorizing  text pieces into predefined classes or categories. For LLMs, text classification demonstrates the model's ability in understanding the content, context, nature, or sentiment of textual data. A powerful language model should accurately assign labels to diverse types of information. The significance of text classification lies in its applications across sentiment analysis, spam detection, and content categorization, contributing to streamlined information organization and retrieval. 

Recently, some researchers compares LLM classified data with human experts. \citet{alizadeh2023open} compare the performance of two widely-used open-source LLMs, HugginChat and FLAN, with that of ChatGPT as well as MTurk on multiple text annotation tasks. Results show that open-source LLMs surpass MTurk in most tasks, and are comparable with ChatGPT in many tasks.
\citet{tornberg2023chatgpt} uses ChatGPT as a classifier for US political messages from Twitter on whether it is from a Democrat or a Republican. Experiment results show that LLM's classified results achieve a better accuracy than the judgements from the Amazon Mturk experts. 

Many works employ LLM augmentation for sentiment analysis. \citet{moller2023prompt} investigate GPT-4 and ChatGPT's ability for generating synthetic data by augmenting small sets of human-generated training samples. Experiments on sentiment analysis and hate speech showcases LLM's great power in data augmentation on classification tasks.
\citet{belal2023leveraging} conduct experiments testing ChatGPT's ability in annotating for sentiment analysis task. Results show that the ChatGPT is a good tool for sentiment analysis, and it has the ability to understand emojis, sarcasm, and irony.

LLM is also proven to be powerful in detecting fake news or human intention.
To improve the results, \citet{smuadu2023fake} conduct data augmentation using GPT-2 to enable domain adaptation on a fake news detection task. The augmentation technique conditionally generates new examples given the news types (left-wing, right-wing, and mainstream).
\citet{fang2023chatgpt} investigate ChatGPT's ability as a data augmentation technique for enhancing compositional generalization
in open intent detection tasks. Experiment results show that ChatGPT's paraphrases for utterances largely outperforms the baseline.

From an application perspective, LLM augmentation is also widely used in medical related classification tasks. 
To address the patient privacy challenge in patient-trial matching task, \citet{yuan2023large} use desensitized patient data as a prompt to guide the ChatGPT in the augmentation process of the trial data. Given the criteria of a clinical trial, the method generates augmented data points adhere to its constraints.
\citet{sarker2023medical} explore ChatGPT for data augmentation in EHR data analysis through prompt engineering. Experiment results show that data augmentation based on ChatGPT improves  performance for both medication identification and medication event classification.

\paragraph{Machine Translation}
Machine Translation is the process of automatically converting text or speech from one language into another. As the LLMs become more powerful, machine translation stands as a critical application, showcasing the model's prowess in comprehending linguistic nuances and producing coherent and contextually accurate translations. The importance of machine translation lies in its ability to bridge language barriers, facilitating global communication and information exchange by rendering content accessible to a wider audience.

For machine translation, data augmentation using LLMs serves two major purposes. The first purpose is enlarge data corpus to faciliate training, and the second purpose is to enhance evaluation quality.
To achieve the first purpose, \citet{oh2023data} explore prompt-based data augmentation approaches that leveraging ChatGPT. It compares three paradigms "paraphrase", "multi-target", and "storytelling" and generates synthetic data on Korean-German language pairs from AI-hub dataset. To achieve the second purpose, DATScore~\cite{eddine2022datscore} utilizes augmented data translated from the source and reference texts using ChatGPT. Experiments on WMT17 and WMT18 shows that DATScore correlates better with human judgements than BLEU, BERTScore~\cite{zhang2019bertscore}, and BARTScore~\cite{yuan2021bartscore} on machine translation task.

\paragraph{Sequence Tagging}
Sequence tagging involves the task of assigning specific labels or tags to individual elements within a sequence, such as words or tokens in a sentence. For LLMs, sequence tagging is a fundamental capability that underscores the model's capacity to discern and annotate various elements in a given context. The significance of sequence tagging lies in its applications across diverse fields, from Named Entity Recognition in natural language processing to part-of-speech tagging in linguistic analysis, enabling more detailed understanding of sequential data.

For sequence tagging, existing works use LLM to generate synthetic data with two major approaches, generating spans and generating labels. 
\begin{itemize}
    \item Generating Spans: \citet{sharma2023and} augment name entity recognition (NER) data by paraphrasing and annotating entity spans in generations using GPT-3 variants. 
    \item Generating Labels: \citet{meoni2023large} propose a weak supervision technique utilizing InstructGPT as a data augmentation tool to predict the annotation. Experiments on a clinical entity extraction showcases the high quality of the augmented data on distilling a Bertbase model. 
    \citet{deng2023holistic} propose a data augmentation method on Universal Decompositional Semantic (UDS) Parsing.  It utilizes ChatGPT to predict the syntactic tree and POS tag and generate pseudo labels for the unlabeled data. 
\end{itemize}

\subsection{Understanding and Inference Tasks} \label{sub:understanding}
In this subsection, we discuss the tasks that are related to text understanding and inference. 
Question Answering tasks demand the models to understand the given question and provide an accurate and relevant answer based on the given context or external knowledge. 
Logical reasoning tasks require the model to apply logical reasoning skills to comprehend and infer from given data or statements.
Natural Language Inference (NLI) tasks involve understanding the relationship between sentences or phrases, such as whether one statement entails, contradicts, or is neutral to another.
LLMs also play an important role in augmenting the datasets for these tasks involved text understanding. 
\paragraph{Question Answering}
Question answering is the cognitive process of providing relevant and accurate responses to posed queries, a task that holds immense importance within LLMs. In this context, question answering demonstrates the model's capacity to understand complex queries, retrieve information from vast datasets, and generate coherent and contextually appropriate answers. The significance of question answering lies in its applications across information retrieval systems, virtual assistants, and educational tools, contributing to more effective and efficient knowledge dissemination.

Question answering task is broad in scope and includes multiple downstream tasks. Some QA tasks are multiple choice based like commonsense reasoning, and fact checking. Other QA tasks are free-form in answer like story completing and machine reading comprehension (MRC). We find that both types of QA tasks can be augmented by large language model.
\begin{itemize}
    \item Commonsense Reasoning: Automate-CoT~\cite{shum2023automatic} proposes a fully automatic pipeline for finding better chain-of-thought prompts to solve commonsense reasoning question answering tasks. This pipeline leverages powerful LLMs like GPT-3 to augment reasoning paths, prune incorrect paths, and select optimal combinations of exemplars.
    \item Story Completing: \citet{whitehouse2023llm} explore the generation of synthetic data for question answering under the multi-lingual setting using various LLMs. This work conducts experiments with dolly-12B, stablevicuna-13B, ChatGPT, and GPT-4 on three datasets that covers over 20 languages on commonsense reasoning and story completing tasks. 
    \item Fact Checking: SCOTT~\cite{wang2023scott} prompts a GPT-neox-20B model to generate annotated data to train a smaller student model to perform QA tasks. Experiments conducted on four QA datasets across commonsense reasoning and fact checking showcase the efficiency of this data augmentation technique. 
    \item MRC: \citet{samuel2023can} introduce a GPT-4 based data augmentation technique targeting the low source setting on machine reading comprehension task. This technique separately generates passsages, questions, and answers using in-context learning. 
    \item MedicalQA: Dr.LLaMA~\cite{guo2023dr} employs ChatGPT and GPT-4 to either rewrite existing medical question-answering pairs or generate new pairs from the training dataset with zero-shot prompting. Such data augmentation helps to train a LLaMA model specialised on medical knowledge.
\end{itemize}

\paragraph{Logical Reasoning}
Logical reasoning involves the ability to analyze and draw conclusions based on given information, a crucial cognitive skill that LLMs aim to master. In the context of LLMs, logical reasoning reflects the model's aptitude for understanding and manipulating symbolic representations, solving problems, and making deductions. The significance of logical reasoning lies in its application across various domains, from puzzle-solving and decision-making to complex problem-solving tasks, showcasing the model's capacity for high-level cognitive functions.

Within the logical reasoning task, Chain-of-Thought has been a popular technique that improves the reasoning process for the language models. CoT is hence frequently used in data augmentation for logical reasoning tasks. 
Fine-tune-CoT~\cite{ho2022large} generates multiple reasoning solutions from LLMs like variants of GPT-3 17B with stochastic sampling to augment the training data for student models. This method largely utilizes the reasoning ability of the LLMs. 
LogiCoT~\cite{liu2023logicot} is a chain-of-thought instruction-tuning dataset designed explicitly for logical reasoning. This dataset is augmented by GPT-4 from high quality logical reasoning data, and is used to fine-tune a LLaMA-7B model. 
Distilling step-by-step~\cite{hsieh2023distilling} is a new simple mechanism for training smaller models with less training data. It prompts PaLM-540B with logical reasoning questions and extracts rationale from it using CoT. 

The reasoning ability of LLMs is proven to be powerful, and the augmented data can be used to improve smaller models or even LLM itself. 
Orca~\cite{mukherjee2023orca} is a 13B model that learns to imitate the reasoning process of larger models. Training data of Orca is augmented from ChatGPT and GPT-4, including explanation traces, step-by-step thought processes, and other complex instructions. 
Instruction backtranslation~\cite{li2023self} uses the LLM to both augment and curate high quality training examples to improve its own performance. Experiments on LLaMA 7B, 33B, and 65B showcase the efficiency of the model on reasoning tasks. \citet{bao2023systematic} propose an evaluation benchmark of LLMs on logical reasoning tasks. It employs logic-driven data augmentation on ChatGPT and GPT-4 to enhance the performance on logical reasoning tasks. The augmentation parts include context, question, and option. 

Data augmentation has also been employed in LLM for science. For instance, \citet{kieser2023educational} use ChatGPT to augment data for physics education, leveraging the LLM's ability to solve quantitative reasoning tasks in physics and concept tests such as the Force Concept Inventory (FCI).

\paragraph{Natural Language Inference}
Natural Language Inference (NLI) is the task of determining the logical relationship between the premise and the hypothesis, such as entailment, contradiction, or neutrality. Within LLMs, NLI represents a sophisticated task that demonstrates the model's ability to understand two statements and infer meaningful connections between them. 

For NLI task, LLMs are powerful in generating augmented data that tackle two challenges. The first challenge is the lack of counterfactual examples, and the second challenge is the out-of-distribution (OOD) or low resouce problem. 
\begin{itemize}
    \item counterfactual example: \citet{li2023large} employ ChatGPT to generate counterfactual samples and counterfactual labels for each original sample. By merging the original data with the counterfactual data, an augmented dataset is formed to enhance the small models' performance on the NLI task. 
    Disco~\cite{chen2023disco} identifies potential locations for performing counterfactual perturbations on the target instances, and then prompts GPT-3 to generate perturbations for NLI task. 
    \item OOD or low resource: \citet{stacey2023improving} generate examples from a GPT-3 model to improve performance in out-of-distribution setting to mimic text that may appear in unseen genres for NLI task. By distilling a Bert model, the performance improves on both MNLI and SNLI settings.
    TDG~\cite{he2023targeted} clusters NLI data into potential challenging subgroups and estimate which subgroups benefit from additional data. Then, it uses GPT-3 coupled with local subgroup models to conduct data augmentation to improve the performance for particularly challenging subgroups.
\end{itemize}

\subsection{Generation and Interaction Tasks} \label{sub:generation}
In this subsection, we discuss the generation and interaction tasks. 
Summarization tasks involve producing a concise and coherent summary of a longer text while retaining the key information and overall meaning.
Data-to-text NLG tasks requires converting structured data into understandable and fluent natural language text.
Open-ended and conditional generation tasks require the model to generate text based on specific conditions or prompts, including creative and unrestricted text generation.
Dialogue tasks involve generating conversational responses in a dialogue setting, requiring the model to maintain context and coherence over multiple turns of conversation.
LLMs can be helpful to not only generate answering but also generate questions to augment the dataset. 
\paragraph{Summarization} 

Text summarization is the process of distilling the most important information from a source text and presenting it in a condensed form. In the realm of Large Language Models (LLMs), summarization becomes a pivotal task, reflecting the model's ability to comprehend, process, and concisely reproduce textual information. The significance of summarization lies in its ability to transform extensive information into a digestible format, aiding in efficient information consumption and understanding. In the context of LLMs, summarization is not just about reducing text length but also about maintaining its essence, coherence, and factual integrity. \citet{chintagunta2021medically} use GPT3 to summarize medical dialogues as labels to train summarization models. \citet{liu2023learning} adopt LLMs (e.g. GPT3D3, ChatGPT, and GPT4) to evaluate the quality (i.e. quality score) of the summarization output by a model, and design a contrastive learning loss function based on the quality score to train a summarization model. \citet{schlegel2023pulsar} use LLMs to generate synthetic data on summarizing patient-doctor dialogues into clinical records. The synthetic data are then used to train the PULSAR model \cite{schlegel2023pulsar}, enabling PULSAR to achieve high scores in the MediQA-Sum task.

% Types of Summarization
Summarization in LLMs can be broadly categorized into two types: extractive and abstractive. Extractive summarization involves identifying key sentences or fragments in the text and stitching them together to form a summary. Here, the LLMs focus on selecting the most informative parts of the original text. In contrast, abstractive summarization requires the LLM to generate new phrases or sentences that capture the core ideas of the text, often paraphrasing or rephrasing the content. This demands a deeper level of understanding and language generation capability from the model. LLMs like GPT-3 and BERT have shown proficiency in both types, with abstractive summarization posing more challenges in terms of generating coherent and contextually accurate summaries.

% Challenges in Summarization
% The primary challenges in LLM-based summarization include maintaining context, managing long documents, and ensuring factual accuracy. Maintaining context involves capturing the essence of the original text without altering its intended meaning, which becomes increasingly difficult with longer documents. Long documents also pose the challenge of identifying which parts are most relevant for the summary. Ensuring factual accuracy is crucial, especially in abstractive summarization, where the model might introduce inaccuracies while paraphrasing. Overcoming these challenges is essential for reliable summarization.

% LLMs in Summarization
% Current LLMs approach summarization tasks using advanced neural network architectures, such as transformer models, which are adept at handling large datasets and complex language patterns. These models are trained on extensive corpora, enabling them to understand and replicate various writing styles and structures. The performance of LLMs in summarization is evaluated based on metrics like ROUGE (Recall-Oriented Understudy for Gisting Evaluation), which measures the overlap between the generated summary and a set of reference summaries. Recent developments have seen LLMs achieve remarkable proficiency, though challenges in coherency and context preservation remain.

\paragraph{Data-to-Text NLG}
Data-to-Text Natural Language Generation (NLG) is a subfield of artificial intelligence that focuses on converting structured data into coherent and readable narrative text. This task is crucial in making data accessible and understandable to a broader audience, transcending the barriers of technical expertise. The significance of Data-to-Text NLG lies in its ability to bridge the gap between raw data and human communication, allowing for effective data-driven storytelling and reporting.

Large Language Models (LLMs) approach Data-to-Text NLG through various methodologies. \citet{krause2023leveraging} design a waterfall
prompting technique using a combination of
both GPT-3 and ChatGPT to enhance Data-to-Text response generation. \citet{li2023autoconv} use LLMs for synthetic conversation generation to tackle the data deficiency problem in training information-seeking conversation models.  

% Large Language Models (LLMs) approach Data-to-Text NLG through various methodologies, each with its unique strengths and applications:
% \begin{itemize}
%     \item Template-Based Methods: These involve predefined templates filled in with data values. While limited in flexibility, they ensure high accuracy and are useful in domains with fixed structures, like weather reports.
%     \item Rule-Based Systems: These systems use a set of linguistic rules to convert data into text. They are more flexible than templates and can generate more varied sentences, but they require extensive domain knowledge to set up.
%     \item Neural Network Models: Modern LLMs predominantly use neural networks, especially transformer-based models, for NLG tasks. These models learn to generate text from large datasets and can produce more fluent and varied narratives. They excel in understanding context and generating text that closely mimics human writing.
% \end{itemize}

% Recent advancements in Data-to-Text NLG include:

% \begin{itemize}
% \item Integration of Multimodal Data: Incorporating not just textual or numeric data but also visual and auditory information to create richer narratives.
% \item Adaptive Learning Models: Developing models that can adapt to different data types and structures without extensive retraining.
% \item Interactivity and User Feedback Incorporation: Creating interactive NLG systems that can refine their output based on user feedback, leading to more accurate and user-centric narratives.
% \end{itemize}

\paragraph{Open-ended and Conditional Generation}
Open-ended and conditional generation in the context of Large Language Models (LLMs) refers to the creation of text that is not strictly predetermined, offering a wide range of possibilities based on given conditions or prompts. This task is particularly challenging for LLMs because it requires a blend of creativity, contextual understanding, and adherence to specified constraints or themes. Open-ended generation pushes the model's capabilities in producing coherent, relevant, and often inventive content, while conditional generation demands adherence to specific guidelines or objectives, adding layers of complexity to the task.

\citet{dai2023chataug} propose AugGPT, which is a text data augmentation approach based on
ChatGPT that rephrases each sentence in the training samples into multiple conceptually similar but semantically different samples. \citet{o2023steering} design Semantic Text Enhancement via Embedding Repositioning (STEER) that utilize LLMs to produce text characteristic of the specific domain. \citet{meadows2023generating} utilize LLMs as a symbolic engine to generate derivations of equations at scale. The augmented data is tested on a finetuned T5 model, showing superior results than GPT models. 

Approaches and Techniques
LLMs employ various models and techniques for open-ended and conditional generation:
\begin{itemize}
    \item Transformer Models: These, like GPT-3, are currently the most prominent in this field. They use deep learning techniques to generate text that is contextually relevant and coherent.
    \item Fine-tuning on Specific Data: This involves training the LLM on specific types of data or genres to improve performance in certain areas, like poetry or technical writing.
    \item Reinforcement Learning from Human Feedback (RLHF): This technique involves training models based on feedback to align the model's outputs more closely with human preferences or specific task requirements.
    
\end{itemize}

% Creative Applications

% In creative domains, LLMs have shown remarkable potential:
% Creative Writing and Poetry: LLMs are used to generate novel ideas, storylines, or even complete drafts in creative writing and to compose poetry with unique styles.
% Artistic Endeavors: Beyond text, these models are being integrated into projects that combine art and technology, like generating scripts for plays or aiding in digital art creation.
% Practical Applications
% The practical applications of open-ended and conditional generation are vast:

% Marketing and Content Creation: LLMs can generate engaging and original content for advertising, social media posts, and blogs.
% Scenario Simulation: In fields like education and training, LLMs can create realistic scenarios for simulations, enhancing learning and practice opportunities.

\paragraph{Dialog}
Dialog systems, a crucial component of human-computer interaction, are designed to enable effective and natural communication between humans and machines. These systems, often manifested as chatbots or virtual assistants, simulate human-like conversation, allowing users to interact with digital devices or services in a more intuitive and familiar manner. The importance of dialog systems lies in their ability to provide efficient, user-friendly interfaces for a variety of applications, from customer service to personal assistance, thereby enhancing user experience and accessibility.

Large Language Models (LLMs) have significantly revolutionized the field of conversational AI. With their advanced natural language processing capabilities, LLMs like GPT-3 have been instrumental in creating chatbots and virtual assistants that can understand and generate human-like text. This advancement has led to more sophisticated and versatile dialog systems capable of handling a wide range of conversational topics and styles. LLMs contribute to conversational AI by improving the fluidity and relevance of interactions, making these systems more engaging and helpful to users.

\citet{zheng2023augesc} use LLMs for dialogue augmentation to support emotional conversation capabilities for dialogue models. \citet{wan2022unified} use T5 as a dialogue user simulator to generate few-shot data augmentations for training conversation models. \citet{li2022controllable} design the Dialogic, which is a dialogue simulation method based on large language models in-context learning to automate dataset creation. Dialogic can automatically select in-context examples for demonstration and prompts GPT-3 to generate new dialogues and annotations in a controllable way, and can further enhance dialogue models' performance. \citet{labruna2023unraveling} utilize ChatGPT to generate goal-oriented dialogues, showing that the ChatGPT annotations are on par with human-generated annotations.

Case Studies and Applications
LLMs have been successfully implemented in various dialog systems, including:
(1) Customer Service: Chatbots powered by LLMs have been employed by numerous businesses to provide instant customer support, handle inquiries, and offer personalized recommendations. \cite{soni2023large,roumeliotis2024llms,pandya2023automating}
(2) Therapy Bots: LLMs are being used in mental health applications, offering conversational therapy and support, though with necessary caution and oversight.\cite{kian2024can,bill2023fine}
(3) Interactive Entertainment: In the gaming and entertainment industry, LLMs have enabled the creation of interactive narratives and characters, enhancing user engagement and experience. \cite{yong2023playing,zhu2023calypso}

\subsection{Multi-Modal Tasks} \label{sub:multimodal}
Other than the pure NLP tasks, multi-modal tasks can also be augmented with LLMs. These tasks either involves generating texts given images or generating images given tasks. Some examples include visual question answering, image captioning, and other tasks. 

\paragraph{Visual Question Answering}
Visual Question Answering (VQA) is a multi-modal task where a system provides answers to questions based on visual content, such as images or videos. This task requires the integration of visual perception with language understanding, challenging the model to interpret visual data and articulate relevant responses in natural language.

SelTDA~\cite{khan2023q} introduces a novel approach for fine-tuning large Visual Language Models (VLMs) on small-scale Visual Question Answering (VQA) datasets by generating question-answer pseudolabels for unlabeled images, effectively augmenting the original dataset. This method enhances model robustness against adversarial questions, improves domain generalization, and retains numerical reasoning skills without needing extra annotations or changes to the architecture. 
T-SciQ~\cite{wang2023t} enhances science question answering by using large language model (LLM) insights to create detailed reasoning paths (CoT) as educational signals. This technique trains smaller models to tackle complex problems with CoT reasoning and introduces a unique data mixing strategy for better learning materials across varying difficulty levels. Demonstrating its effectiveness, T-SciQ sets a new record on the ScienceQA benchmark with a 96.18\% accuracy, outdoing the top fine-tuned baseline by 4.5\%.

To tackle Visual Question Answering (VQA) in scenarios with limited labeled data, \citet{askarian2022inductive} expand the initial dataset, infusing the VQA model with enhanced inductive biases through newly generated questions based on image annotations. This approach yields up to a 34\% accuracy improvement over baseline models trained solely on the original labeled dataset.

VQA has practical applications in areas such as accessibility technology, where it aids visually impaired users in understanding their surroundings. It's also used in customer service to answer queries based on product images and in educational tools that help students learn through visual aids.

\paragraph{Image Captioning and Editing}
Image captioning is a multi-modal task that involves generating descriptive text for images. This requires a system to not only recognize the visual elements within an image but also to understand their context and how they relate to each other, synthesizing this information into coherent, natural language sentences. It bridges the gap between visual perception and language generation, challenging models to accurately interpret visual data and express these interpretations as human-like captions.

ChatBridge~\cite{zhao2023chatbridge} connects text, images, videos, and audio using language.  It leverages LLM and extends their zero-shot capabilities to incorporate diverse multi-modal inputs. The training includes two stages, first learning to link each modality with language, then fine-tuning with the dataset MULTIS for specific multi-modal tasks. 
\citet{xiao2023multimodal} utilize Stable Diffusion to generate high-quality image-caption pairs for multimodal data augmentation, showing significant improvements on the MS COCO dataset, especially with limited training data. This approach surpasses previous unpaired image captioning methods and further enhances training efficiency and effectiveness through quality-based filtering of generated data. 
ALIA~\cite{dunlap2024diversify} leverages vision and language models for automated language-guided image editing, enriching training datasets while preserving class-relevant information. This method enhances dataset diversity without compromising visual consistency, outperforming traditional and text-to-image augmentation in fine-grained classification tasks, improving domain generalization and reducing contextual bias. 

Image captioning has diverse applications in various fields, enhancing accessibility for visually impaired users by providing textual descriptions of images on the web or in digital media. It plays a crucial role in social media, where automatic captioning can improve user engagement and content accessibility. Furthermore, in the educational sector, image captioning aids in creating more interactive and accessible learning materials, allowing students to gain a deeper understanding of visual content. 

\paragraph{Other multi-modal tasks}
Other than visual question answering and image captioning, some other multi-modal tasks also use LLM to conduct data augmentation. One example is Multi-modal Named Entity and Relation Extraction. \citet{chen2023chain} distill the reasoning ability of LLMs into small student model by generating a intermediate reasoning steps (CoT). It draws out reasoning skills from LLMs using detailed prompts that include multi-grain (noun, sentence, multi-modality) and data augmentation (style, entity, image) dimensions. The new technique simplifies the complex reasoning into a form that smaller models can use. Another example is tooling. 
GPT4Tools~\cite{yang2024gpt4tools} employs self-instruct to enable open-source LLMs to use multi-modal tools. We will explain this paper more in detail in the next sub-section.

\subsection{Other Tasks} \label{sub:others}
Other than the above mentioned tasks, many other NLP tasks also employ LLMs as data augmenting tools. We will discuss them in this subsection. 

\paragraph{Tooling}
Tooling tasks in the context of Large Language Models (LLMs) refer to the application of these models in developing tools that aid in various aspects of programming, content creation, and data analysis. These tasks leverage the language understanding and generation capabilities of LLMs to automate, optimize, and enhance various workflows and processes.

One key area of data augmentation's usage of tooling is in code generation and assistance.  LLMs are employed to assist in writing and optimizing code, offering suggestions, debugging help, and even generating code snippets based on user requirements. \citet{dong2023boosting} explore the existing data augmentation techniques from NLP and graph learning for enhancing training quality in source code learning. After reviewing and categorizing the literature, 11 data augmentation methods from NLP and graph learning are identified as potentially applicable to source code learning. \citet{zhang2023toolcoder} enhance code generation by integrating tool usage information via an automated annotation method and API search tools, showing significant improvements in benchmarks. Despite its smaller size, ToolCoder's performance is comparable to GPT-3.5, illustrating the benefits of embedding programming tools into code generation models.

Another possible area of data augmentation using LLM is to train models as classification or detection tools. For instance, \cite{smuadu2023fake} work as very good hyper-partisan news detection tool. It employs GPT-2 employ to conditionally generate new examples given the news types (i.e., left-wing, right-wing, and mainstream). Then, it fine-tunes the GPT-2 base model on the hyperpartisan Buzzfeed dataset to generate new samples. These detection models are very useful in industries and applications. 

Other than the above-mentioned tooling tasks, LLM can also be used to augment data for setting up a comprehensive benchmark for multiple tools. GPT4Tools~\cite{yang2024gpt4tools} empowers open-source large language models like LLaMA and OPT with tool-using abilities, leveraging Low-Rank Adaptation (LoRA) optimization for visual tasks using GPT-4 for augmentation. It includes a benchmark for evaluating tool use in models, showing significant improvements in both seen and unseen tool invocation accuracy through extensive experiments.

% Challenges and Solutions
The main challenges in tooling with LLMs include ensuring accuracy, maintaining context relevance, and integrating seamlessly with existing tools and workflows. Solutions involve continuous model training, user feedback loops for model refinement, and developing intuitive interfaces that facilitate easy integration and usage.

\section{Domains}
\label{sec:domains}
The preceding sections have outlined prominent data augmentation techniques applied across diverse tasks. In this section, we delve further into categorizing the specific application domains of these data augmentation methodologies. We systematically categorize these domains into distinct areas: clinical, finance, legal and social science.

\subsection{Clinical}
%TODO: polish entire subsection

%1. LLMs have remarkable abilities and can be beneficial for clinical application such as xxx.  
%2. Directing applying LLMs faces challenges where one of the most tobale is data scaracity.
%3. A number of work focus on data augmentation to facilate the research on LLMs in clinical domain.
%4. follow by an enumeration of work and citations..

The emergence of clinical applications powered by LLMs is poised to transform the healthcare system. LLMs such as GPT-4 have demonstrated a remarkable ability to understand and generate text, making them valuable tools for tasks ranging from enhancing medical transcription to assisting with medical diagnoses. Integrating these models into clinical settings can significantly enhance the efficiency and accuracy of various healthcare processes, but this potential is accompanied by major challenges. For example, establishing close alignment of model predictions with the assessment of medical professionals and addressing biases inherent in the training data must be addressed. Navigating these challenges is essential for realizing the full potential of large language models in healthcare and ultimately advancing the quality of patient care and medical research. 

One of the major obstacles to adapting LLMs for clinical tasks is the scarcity of data. First, the accessibility of medical data is limited. In the United States, for example, the distribution of medical data must abide by HIPAAA regulations, which protect patient information. In addition, annotation of medical data is a time-intensive and costly undertaking~\cite{chintagunta2021medically, meoni2023large}, further restricting the availability of data for AI development in the clinical domain. Together, these limitations necessitate alternative routes to generating large, high-quality datasets. 

To this end, an emerging stream of research is focusing on data augmentation approaches that generate synthetic data to facilitate the adaptation of LLMs in medical domain. Electronic Health Records (EHRs) play a crucial role in the modern clinical field. Several works propose data augmentation approaches to facilitate the usage of LLMs for EHR-related applications. \citet{meoni2023large} explore annotating EHR data with InstructGPT-3 to improve the performance of distilled BERT-based models for multilingual clinical entity extraction. The proposed method outperforms dictionary supervision in extracting clinical entities on the E3C multilingual data set \cite{magnini2022european}. \citet{sarker2023medical} tackles the medication identification and event classification problem in EHRs. By prompting ChatGPT to rephrase original sentences of the Contextualized Medication Event Dataset, the authors enhance the performance of different pre-trained BERT models. \citet{yuan2023large} prompt LLMs with desensitized patient data to create privacy-aware supplementary data points to comprehend EHRs for clinical patient-trial matching. \citet{guevara2024large} utilize LM-generated synthetic clinical text at the fine-tuning stage to improve LMs ability on extracting social determinants of health, such as employment status and housing issues, from EHRs. Aside from prompting LLMs to obtain synthetic data, recent work explores generating mixed-type tabular EHRs with diffusion models~\cite{ceritli2023synthesizing}.

Medical question answering is another field of interest. \citet{guo2023dr} develop smaller efficient language models with augmentation data generated by LLMs on the PubMedQA dataset. The proposed method uses GPT-3.5 Turbo and GPT-4 to rewrite existing QA pairs or generate new pairs via zero-shot prompting. Results show that the small fine-tuned model with the data augmentation approach outperforms few-shot GPT4 on the PubMedQA dataset. For medical dialogue applications, GPT-3-ENS is introduced to create high-quality synthetic training data via few-shot learning and an ensemble method. Together with the human labeled data, the generated data help yield more performant models in terms of medical accuracy and coherency \cite{chintagunta2021medically}.

% \cite{yang2023bliam}prompt PubMedBERT with a manually constructed list of cloze promptstriplets (drug A, drug B, cell line) + label. In addition to the promising prediction performance, the data points synthesized by BLIAM are interpretable and model-agnostic, enabling in silico augmentation for in vitro experiments.

\subsection{Finance}
One major line of research centers on leveraging LLM-based data augmentation for finance-related reasoning tasks. To improve the numerical reasoning ability for financial question answering tasks, \citet{hwang2023augmentation} put forth a novel context augmentation method that generates synthetic financial contexts based on the arithmetic program and operators present in the question. To address the limited availability of labeled datasets in financial analysis and interpretation tasks, \citet{chu2023datacentric} propose an abductive augmentation reasoning framework that refines the pseudo-labels generated from a small labeled dataset. This aims to correct noisy labels and improve the quality of training data to enhance the performance of the financial large language model.

Another application is intent detection, that aims to classify the user’s intent given an utterance. To alleviate the problem of data scarcity, researchers employ LLM-based data augmentation techniques to generate synthetic training data. This involves creating artificial utterances corresponding to given intents. A common strategy is to construct a data generation and filtering pipeline. \citet{sahu-etal-2022-data} prompt GPT models to augment the training data for each intent class. Subsequently, the same GPT model is utilized as a classifier to filter and relabel the generated data. Similarly, \citet{lin-etal-2023-selective} employ GPT-3 and OPT to synthesize new data points for each class, then apply pointwise V-information to filter out examples that are not relevant to the desired intent. Both methods achieve state-of-the-art performance on Banking77 \citep{casanueva-etal-2020-efficient}, a challenging intent detection dataset with fine-grained banking-related intents.

\subsection{E-commerce}

% Another prospective application of applying LLMs data augmentations is the recommender systems. 

Adopting data augmentation techniques by LLMs in E-commerce applications such as search and recommender systems, shopping assistance, and customer supports are prospective trends in the development of LLMs.

Recommender systems are algorithms designed to predict and suggest items of interest to users, based on their preferences and behavior \cite{guo2020survey,gao2023survey,recommender-gnn,hao2021re,zhou2020improving,da2020recommendation}. These systems play a crucial role in various domains, such as e-commerce, entertainment, and content platforms, by personalizing user experiences and enhancing engagement. The combination of recommender systems with LLMs is an emerging trend where LLMs can largely enhance the capability of the recommender systems \cite{fan2023recommender}. \citet{cui2022m6,christakopoulou2023large,hua2023up5} utilize LLMs with prompt tuning: by adding prompt tokens to LLMs and then updating them based on task-specific recommendation datasets. 

\subsection{Social Sciences}
Large language model (LLM)-based data augmentation provides the potential for advancing scientific understanding of human social and political behavior. Several studies have demonstrated that LLMs serve as effective surrogates for human respondents across various social science tasks. For instance, \citet{törnberg2023chatgpt4} use ChatGPT to annotate and classify Twitter messages, showcasing the model's ability to accurately identify political affiliations by reasoning based on contextual knowledge and inferences about the author's intentions, traditionally considered uniquely human qualities. In addition, another study \citep{article} reveal that GPT-3 can simulate viewpoints of demographically diverse sub-populations within the U.S. political domain. Conditioning an LLM with socio-demographic profiles allows it to mirror responses of humans with matching demographic traits, uncovering patterns between ideas, attitudes, political views, and socio-cultural contexts. This suggests that synthetic samples generated by LLMs offer a reliable means for researchers to explore hypotheses before deploying costly studies involving human subjects in social science research. \citet{griffin-etal-2023-large} extend the scope beyond static psychological modeling, illustrating LLMs' ability to model dynamic psychological changes in response to influencing input.

The capability of LLMs in generating human-like responses extends their application in data augmentation to education research. \citet{kieser2023educational} delve into the potential of employing ChatGPT to synthesise data that closely resemble students from various cohorts possessing different preconceptions. These data augmentation techniques could be of interest to education researchers as it would save a significant amount of time and effort in the development of assessments and concept tests.

% LLMs can be applied to social network analysis, with their strong capability of modeling complex data and generating useful information for social network analysis.

LLMs can enhance social network analysis due to their adeptness in modeling complex data and extracting valuable insights \cite{zeng2024large}. Social networks, both online and offline, serve as fundamental structures for human interaction and information dissemination. These networks encapsulate the intricate web of relationships between individuals, organizations, and communities, forming the basis of societal dynamics. Computational approaches leverage techniques such as graph theory, graph representation learning, and data mining to simulate, infer, or construct social networks from large-scale data sources \cite{jo2022score,luo2023fast,mo2023conditional,mo2024conditional}. These generated networks not only provide insights into human behavior and social dynamics but also facilitate various applications such as recommendation systems, targeted advertising, and sentiment analysis.

LLMs have emerged as powerful tools for generating social knowledge graphs due to their advanced natural language processing capabilities. By analyzing vast amounts of textual data from diverse sources such as social media, forums, and news articles, LLMs can extract valuable information about relationships, entities, and events. Through their ability to understand context, infer semantic meaning, and detect patterns, LLMs can construct comprehensive and accurate representations of social networks. These generated knowledge graphs not only provide insights into the structure and dynamics of social interactions but also facilitate tasks such as community detection, trend analysis, and information retrieval. Additionally, LLMs offer the flexibility to adapt and evolve with changing data and can continuously refine and update social knowledge graphs to reflect real-world dynamics.

% Giving a set of clear and concise instructions through prompts, expensive LLMs, such as GPT-4, can be used as near-human annotators to create a collection of labeled data across different domains. 

Another emergent direction of research is focused on the field of psychology. \citet{neuman2023data} contribute to this area with a personality data augmentation approach that combines LLMs and domain expertise. Their method involves training GPT-2 to generate specific personality types by completing sentences carefully selected by domain experts representing the personality's beliefs. This approach proves particularly valuable in personality modeling tasks where there is a scarcity of large amounts of high-quality labeled data, such as labeled texts from clinically diagnosed psychopaths.
\citet{liyanage-etal-2023-augmenting}
focus on the challenge of imbalanced datasets in mental wellness on Reddit posts. They design prompt-based data augmentation methods using ChatGPT and GPT-3 models to aid in the development of classification models for detecting mental health issues. 
Meanwhile, \citet{zhang-etal-2023-ask} propose enhancing dialogue models for mental health support through data augmentation with additional annotations that take the form of reasoning support provided by prompt-based interactions with LLM experts, aiming to improve the robustness of the models in addressing mental health concerns.

% Deception Detection

% - Data Augmentation for Fake Reviews Detection

% - People Make Better Edits: Measuring the Efficacy of LLM-Generated Counterfactually Augmented Data for Harmful Language Detection

% Hate Speech Detection

% - Generation-Based Data Augmentation for Offensive Language Detection: Is It Worth It?

\subsection{Legal}
There is a rising interest in incorporating LLMs into the legal domain to automate tasks such as analysis and generation of legal documents. DALE is introduced to generate coherent and diverse augmentations augmentation data for low-resource legal NLP tasks \cite{dale}. DALE uses selective masking and conditional generation based on a BART model on legal documents. The generated data is then utilized to enhance performance of downstream models in low-resource Legal NLP tasks. \citet{contractdraft} proposes integrating traditional AI and generative AI techniques, applying ChatGPT for contract drafting, and evaluating AI-generated clauses with LLMs to streamline the task for legal professionals.

\newpage

\end{document}